\documentclass[runningheads]{llncs}

\usepackage{todonotes}

\usepackage[T1]{fontenc}
\usepackage{graphicx}

\usepackage{amsfonts}
\usepackage{amsmath}
\usepackage{amssymb}
\usepackage{mathtools}
\usepackage{booktabs}
\usepackage{comment}
\usepackage{xspace}
\usepackage{nicefrac}

\usepackage{algorithm}
\usepackage{algorithmic}
\usepackage{subcaption}

\usepackage[noabbrev,nameinlink]{cleveref}

\usepackage{tikz}
\usetikzlibrary{patterns,arrows,arrows.meta,calc,shapes,shadows,decorations.pathmorphing,decorations.pathreplacing,automata,shapes.multipart,positioning,shapes.geometric,fit,circuits,trees,shapes.gates.logic.US,fit,backgrounds,decorations.text,topaths}

\newcommand{\ie}{\emph{i.e.}\@\xspace}
\newcommand{\eg}{\emph{e.g.}\@\xspace}

\newcommand{\Storm}{\textsc{Storm}\@\xspace}
\newcommand{\Prism}{\textsc{Prism}\@\xspace}
\newcommand{\Prophesy}{\textsc{PROPhESY}\@\xspace}

\DeclareMathOperator*{\argmax}{argmax}

\newcommand{\Lone}{$L_1$}
\newcommand{\parto}{\rightharpoonup}

\newcommand{\bN}{\mathbb{N}}

\newcommand{\bR}{\mathbb{R}}

\newcommand{\bP}{\mathbb{P}}

\newcommand{\cD}{\mathcal{D}}

\newcommand{\cP}{\mathcal{P}}
\newcommand{\cU}{\mathcal{U}}

\newcommand{\sP}{\mathsf{P}}
\newcommand{\sNP}{\mathsf{NP}}
\newcommand{\sPSPACE}{\mathsf{PSPACE}}
\newcommand{\sEXP}{\mathsf{EXPTIME}}
\newcommand{\sETR}{\mathsf{ETR}}

\newcommand{\real}{\bR}

\newcommand{\dist}{\cD}

\newcommand{\path}{\tau}
\newcommand{\last}{\mathit{last}}
\newcommand{\states}{\mathit{states}}
\newcommand{\paths}{\mathsf{Paths}}

\newcommand{\policy}{\pi}

\newcommand{\discount}{\gamma}

\newcommand{\sinit}{s_\iota}

\newcommand{\Vopt}{\overline{V}}
\newcommand{\Vpes}{\underline{V}}

\newcommand{\Qpes}{\underline{Q}}
\newcommand{\policyopt}{\overline{\policy}}
\newcommand{\policypes}{\underline{\policy}}

\newcommand{\Plow}{\check{P}}
\newcommand{\Pup}{\hat{P}}

\newcommand{\Pest}{\tilde{P}}

\DeclarePairedDelimiter\tup{\langle}{\rangle}

\begin{document}
\title{Robust Markov Decision Processes:\\A Place Where AI and Formal Methods Meet}
\titlerunning{Robust MDPs: A Place Where AI and Formal Methods Meet}
\author{Marnix Suilen\inst{1} %
\and
Thom Badings\inst{1}
\and
Eline M. Bovy\inst{1} %
\and\\
David Parker\inst{2} %
\and
Nils Jansen\inst{1,3} %
}
\authorrunning{M. Suilen et al.}
\institute{Radboud University, Nijmegen, The Netherlands \and
University of Oxford, United Kingdom \and
Ruhr-University Bochum, Germany
}
\maketitle             

\begin{abstract}
Markov decision processes (MDPs) are a standard model for sequential decision-making problems and are widely used across many scientific areas, including formal methods and artificial intelligence (AI).
MDPs do, however, come with the restrictive assumption that the transition probabilities need to be precisely known.
\emph{Robust} MDPs (RMDPs) overcome this assumption by instead defining the transition probabilities to belong to some \emph{uncertainty set}.
We present a gentle survey on RMDPs, providing a tutorial covering their fundamentals.
In particular, we discuss RMDP semantics and how to solve them by extending standard MDP methods such as value iteration and policy iteration.
We also discuss how RMDPs relate to other models and how they are used in several contexts, including reinforcement learning and abstraction techniques. 
We conclude with some challenges for future work on RMDPs.

\end{abstract}

\section{Introduction}

Markov decision processes (MDPs) are a fundamental model for tackling decision making under uncertainty across various areas, such as formal methods~\cite{DBLP:conf/lics/Katoen16}, operations research~\cite{DBLP:books/wi/Puterman94}, and artificial intelligence~\cite{DBLP:books/lib/SuttonB98}.
At the core of MDPs is the assumption that the transition probabilities are precisely known, a requirement that is often prohibitive in practice~\cite{DBLP:journals/sttt/BadingsSSJ23}.
For example, in data-driven applications in AI such as reinforcement learning (RL)~\cite{DBLP:books/lib/SuttonB98}, transition probabilities are unknown and can only be estimated from data.
Furthermore, in formal verification problems, the state-explosion problem often prevents the model from being fully built~\cite{DBLP:series/lncs/BaierHK19,DBLP:books/daglib/0020348,DBLP:conf/laser/ClarkeKNZ11}. 
As a remedy, sampling-based approaches that only estimate the transition probabilities, such as statistical model checking~\cite{DBLP:conf/cav/AshokKW19,DBLP:conf/rv/LegayDB10}, are used.
Any sampling-based approach naturally carries the risk of statistical errors and hence, incorrect estimates of the probabilities.

\emph{Robust} MDPs (RMDPs) overcome this assumption of precise knowledge of the probabilities.
An RMDP contains an \emph{uncertainty set} that captures all possible transition functions from which an adversary, typically called \emph{nature}, may choose.
Tracing back to at least interval Markov chains in the formal methods community~\cite{DBLP:conf/lics/JonssonL91} and bounded-parameter MDPs in AI~\cite{DBLP:journals/ai/GivanLD00}, RMDPs provide a general and flexible framework for modelling MDPs with uncertainty on the transition probabilities~\cite{DBLP:journals/mor/Iyengar05,DBLP:journals/ior/NilimG05,DBLP:journals/mor/WiesemannKR13}.
RMDPs provide a rigorous approach to quantifying the impact of data-driven methods for MDPs and, as such, represent an important topic in the intersection of AI and formal methods.

\subsubsection{RMDPs in AI and formal methods.}
As RMDPs are studied across several research fields, results inevitably become scattered across the communities.
While several recent algorithmic developments and applications of RMDPs stem from AI and operations research, see \eg,~\cite{DBLP:journals/jair/BadingsRAPPSJ23,DBLP:journals/mor/GoyalG23,DBLP:journals/jmlr/HoPW21} and many of the other works cited in this survey, tool support is arguably more mature in the formal methods community. %
While several of the major contributions to dynamic programming for RMDPs can be traced back more to the AI than the formal methods community, these algorithms have been implemented in probabilistic model checkers such as \Prism~\cite{DBLP:conf/cav/KwiatkowskaNP11} and \Storm~\cite{DBLP:journals/sttt/HenselJKQV22}, which are well-known within formal methods but less so in AI.
Because work on RMDPs in formal methods and AI faces many of the same problems, we believe that research in both communities can benefit greatly from each other.
In particular, theoretical contributions in one field may improve tool support in the other.
Vice versa, improved tool support may lead to more advanced applications of RMDPs across both research areas.

\subsubsection{Goal of this survey.}
The goal of this survey is to unify the views on RMDPs from the AI and formal methods communities.
While the theory of RMDPs has made significant advances over the years, surveys summarizing these results are as of yet sparse.
To the best of our knowledge,~\cite{ou2024sequential} is the only other survey on RMDPs available, with a primary focus on summarizing recent technical results.
In contrast, this paper aims to provide an introduction to the theory of robust MDPs and a short review of its connections with other well-known models and applications in the areas of formal methods and AI.

\subsubsection{Outline.}
This survey consists of two main parts.
In the first part, we review the basics of MDPs in \Cref{sec:prelims} and then lay out the theoretical foundations underpinning RMDPs in \Cref{sec:robust:mdps}.
These sections are meant to be accessible to readers with basic familiarity with MDPs.
In the second part, we provide a gentle survey of the existing literature, in particular focusing on connections with other models (\Cref{sec:connections}) and applications and tool support (\Cref{sec:applications}).
We conclude in \Cref{sec:challenges} with some interesting directions for future work, both in theory and applications, for RMDPs.

\section{Markov Decision Processes and How to Solve Them}\label{sec:prelims}

For a set $X$, we write $|X|$ for its cardinality.
Partial functions are denoted by $f \colon X \parto Y$, and we write $\bot$ for \emph{undefined}.
A \emph{discrete probability distribution} over a finite set $X$ is a function $\mu \colon X \to [0,1]$ such that $\sum_{x \in X} \mu(x) = 1$.
The set of all probability distributions over $X$ is denoted by $\dist(X)$.
A distribution is called Dirac if it assigns probability one to precisely one element and zero to all others.

\subsection{Markov Decision Processes}
We define Markov decision processes (MDPs) and their semantics.

\begin{definition}[MDP]
    A Markov decision process (MDP) is a tuple of the form $(S,\sinit,A,P,R)$, where $S$ is a finite set of states with $\sinit \in S$ the initial state, $A$ is a finite set of actions, $P \colon S \times A \parto \dist(S)$ is the probabilistic transition function, and $R \colon S \times A \parto \real_{\geq 0}$ is the (non-negative) reward function.
\end{definition}

We focus on expected reward objectives and hence omit a labelling function from our MDPs.
We use partial functions for the transition and reward functions to allow for enabled actions.
An action is enabled if $P(s,a)$ is defined.
We write $A(s) \subseteq A$ for the set of enabled actions at state $s$.
We require that the transition and reward function are consistent with each other, that is, $P(s,a) = \bot \iff R(s,a) = \bot$.
For convenience, we write $P(s,a,s')$ for the probability $P(s,a)(s')$.

A path in an MDP is an (in)finite sequence of successive states and actions: $\path = (s_0,a_0,s_1,\dots) \in (S \times A)^* \times S$ where $s_0 = \sinit$ and $\forall i \in \bN,\;P(s_i,a_i,s_{i+1}) > 0$.
A path is finite if the sequence is finite, $\path = (s_0,a_0,\dots,s_k)$, for which we write $\last(\path) = s_k$ for the last state.
The set of all paths is denoted as $\paths$.
The sequence of states in a path $\tau = (s_0,a_0,s_1,\dots)$ is $\states(\tau) = (s_0,s_1,\dots)$.

A \emph{discrete-time Markov chain} (DTMC) is an MDP with only one enabled action in each state: $\forall s \in S,\; |A(s)| = 1$.
For DTMCs, we omit the actions altogether from the tuple and write $(S,\sinit,P,R)$.

A \emph{policy} (also called \emph{scheduler} or \emph{strategy}) is a function that maps paths to distributions over actions $\policy \colon \paths \to \dist(A)$.
Such policies are called \emph{history-based} and \emph{randomized}.
A policy is \emph{deterministic} if it only maps to Dirac distributions over actions and \emph{stationary} (also called \emph{memoryless}) if it only considers finite paths of length one.
Stationary deterministic (also called positional) policies are written as $\policy \colon S \to A$. 

Given a policy $\policy$ for an MDP $M$, the action choices in $M$ are resolved, resulting in a DTMC.
\begin{definition}[Induced DTMC]\label{def:induced:DTMC}
    Let $M = (S,\sinit,A,P,R)$ be an MDP and $\policy \colon \paths \to \dist(A)$ a policy.
    The induced DTMC is a tuple $M_\policy = (S^*,\sinit,P_\policy,R_\policy)$, where $S^*$ is the (infinite) set of states, $\sinit$ is the initial state, and the transition and reward functions are defined as
    \begin{align*}
    P_\policy(\states(\path),\states(\path) : s') &= \sum_{a \in A} \policy(\path)(a) \cdot P(\last(\tau),a,s'),\\
    R_\policy(\states(\path)) &= \sum_{a \in A} \policy(\path)(a) \cdot R(\last(\path),a),
    \end{align*}
    where $\path \in \paths$ and $\states(\path) : s'$ denotes concatenation of $\states(\path)$ with $s'$.
\end{definition}
The induced DTMC $M_\policy$ has a unique probability measure $\bP_{M_\policy}$ that follows from the standard cylinder set construction, see, \eg,~\cite{DBLP:books/daglib/0020348,DBLP:journals/corr/abs-2305-10546}.
For stationary policies, the set of states of the induced DTMC coincides with that of the MDP and is thus finite.

\subsubsection{Objectives.}
The primary objective we consider in this paper is cumulative reward maximization until reaching a state in some target set $T \subseteq S$~\cite{DBLP:conf/sfm/KwiatkowskaNP07}.
We shall simply call this objective \emph{reach-reward}.
The goal is to compute the expected reward and an associated optimal policy for this objective in a given MDP $M$, which we formalize in the following subsection.
Other common objectives include \emph{reachability}, \emph{discounted expected reward}, \emph{reach-avoid}, and general temporal logic objectives expressed in (probabilistic) LTL~\cite{DBLP:conf/focs/Pnueli77} or CTL~\cite{DBLP:journals/fac/HanssonJ94}.

\subsection{Classical Dynamic Programming}

We review dynamic programming for MDPs with an infinite horizon undiscounted reward objective in preparation for our discussion of \emph{robust} dynamic programming in \Cref{sec:robust:mdps}.
We define state and state-action value functions $V \colon S \to \real$ and $Q \colon S \times A \to \real$, respectively.
Dynamic programming updates these value functions iteratively until the least fixed point is reached.

We preprocess the set of states $S$ based on graph properties via the following standard procedure~\cite{DBLP:books/daglib/0020348} and \Prism-semantics for reward objectives~\cite{FKNP11,DBLP:conf/sfm/KwiatkowskaNP07}.
Let $T \subseteq S$ be the target set of our objective, let $S^{\infty} \subseteq S$ be the set of states for which there exists a policy that does not reach $T$ almost-surely, and denote the remaining states by $S^? = S \setminus (T \cup S^\infty)$.
For all $a \in A$ and target states $s \in T$, let $Q(s,a) = 0$.
Similarly, for all $a \in A$ and $s \in S^\infty$, let $Q(s,a) = \infty$.
For all other states $s \in S^?$ and $a \in A$, we initialize the state-action values as $Q(s,a) = 0$.
We iteratively update the state and state-action values for all $(s,a) \in S^? \times A$ by:
\begin{align*}
 V^{(n)}(s) = \max_{a \in A} Q^{(n)}(s,a), \quad Q^{(n+1)}(s,a) = R(s,a) + \sum_{s' \in S} P(s,a,s') V^{(n)}(s').
\end{align*}
This process is also known as \emph{value iteration}.
When performing value iteration, we do not need to keep track of the state-action values $Q$ explicitly but instead can directly compute the state-values $V$ by setting $V(s) = 0$ for all $s \in T$, for all $s \in S^\infty, \, V(s) = \infty$, and for all $s \in S^?$ we iteratively compute:
\begin{align}
V^{(n+1)}(s) =  \max_{a \in A} \left\{ R(s,a) + \sum_{s' \in S} P(s,a,s')V^{(n)}(s') \right\},\label{eq:bellman:mdp}
\end{align}
The optimal value function $V^*$ is the unique least fixed point of the \emph{Bellman equation} in~\Cref{eq:bellman:mdp}. 

For many objectives in MDPs, such as reach-reward maximization, optimal policies are stationary and deterministic, \ie, of type $\policy \colon S \to A$~\cite{DBLP:books/wi/Puterman94}. 
An optimal stationary deterministic policy $\policy^*$ that achieves value $V^*$ can be extracted by performing the following one-step dynamic programming procedure:
\[
\policy^*(s) = \argmax_{a \in A} \left\{ R(s,a) +  \sum_{s' \in S} P(s,a,s')V^*(s') \right\}.
\]

\subsubsection{Policy evaluation and improvement.}

As an alternative to value iteration, MDPs can also be solved through \emph{policy iteration}.
Policy iteration consists of two alternating steps: \emph{policy evaluation} and \emph{policy improvement}.
Policy evaluation is the process of computing the value of an MDP for a given policy, which is also known as verifying or model checking the induced Markov chain from \Cref{def:induced:DTMC}~\cite{DBLP:books/daglib/0020348}.
The value of a stationary policy $\policy \colon S \to \dist(A)$ is computed by the following Bellman equation:
\[
V^{(n+1)}_\policy(s) = \sum_{a \in A} \policy(s,a) \cdot \left(R(s,a) +  \sum_{s' \in S} P(s,a,s') V^{(n)}_\policy(s') \right).
\]
Alternatively, we may explicitly construct the induced DTMC $(S,\sinit, P_\policy, R_\policy)$ from \Cref{def:induced:DTMC}, whose set of states coincides with that of the MDP (and is thus finite) as the policy $\policy$ is stationary.

After evaluating the current policy $\policy$ and determining its value function $V^*_\policy$, the \emph{policy improvement} step looks for a new policy $\policy'$ that outperforms the current policy as follows.
First, compute the state-action values under $\policy$ as
\[
Q_{\policy}(s,a) = R(s,a) +  \sum_{s'} P(s,a,s')V^*_\policy(s'), \quad \forall s \in S, \, a \in A(s).
\]
The new policy $\policy'$ is extracted as $\policy'(s) = \argmax_{a \in A} Q_\policy(s,a)$ for all $s \in S$ and has a value at least as good as the previous policy, \ie, $V^*_{\policy'} \geq V^*_{\policy}$.
This process terminates as soon as the policy does not change anymore: $\policy' = \policy$, after which $\policy$ is guaranteed to be optimal.

\subsubsection{Modifications towards other objectives. }
Many other objectives, such as reachability and discounted reward, can be solved by straightforward modifications to the Bellman equation.
For maximizing the reachability probability of a target set $T \subseteq S$, the reward function is removed, and the preprocessing step is changed to set $Q(s,a) = 1$ for all $(s,a) \in T \times A$, and $Q(s,a) = 0$ for all $(s,a) \in S^{\infty} \times A$.
For discounted reward, the preprocessing is removed altogether, and all state-action pairs are initialized with $Q(s,a) = 0$.
The Bellman equation from \Cref{eq:bellman:mdp} is modified for both cases, respectively:
\begin{align}
    Q^{(n+1)}(s,a) &= \sum_{s' \in S} P(s,a,s') V^{(n)}(s'), \tag{\text{reachability}} \\
    Q^{(n+1)}(s,a) &= R(s,a) + \discount \sum_{s' \in S} P(s,a,s') V^{(n)}(s') \tag{discounted}.
\end{align}
These modifications can also be directly applied to the state-value function $V$ from \Cref{eq:bellman:mdp}.

\subsubsection{Variations and other methods.}
Several variations to value iteration have been introduced to resolve issues with accuracy and convergence, such as
\emph{bounded value iteration}~\cite{DBLP:conf/atva/BrazdilCCFKKPU14,DBLP:conf/rp/HaddadM14}, \emph{interval iteration}~\cite{DBLP:conf/cav/Baier0L0W17}
\emph{optimistic value iteration}~\cite{DBLP:conf/cav/HartmannsK20} and \emph{sound value iteration}~\cite{DBLP:conf/cav/QuatmannK18}.
As an alternative to dynamic programming, MDPs can also be naturally encoded as a linear optimization problem which can be solved in polynomial time for many objectives (among which: reach-reward, discounted reward, reachability)~\cite{DBLP:books/daglib/0020348}.
An extensive experimental evaluation comparing several methods for solving MDPs can be found in~\cite{DBLP:conf/tacas/HartmannsJQW23}.

\section{Theory of Robust Markov Decision Processes}\label{sec:robust:mdps}
Having briefly recapped the basics of MDPs and dynamic programming, we now move to \emph{robust} MDPs.
In the following, let $X$ be a set of variables.
An \emph{uncertainty set} $\cU$ is a non-empty set of variable assignments subject to some constraints and is defined as $\cU = \{f \colon X \to \bR \mid \text{constraints on } f\}$.

\begin{definition}[RMDP]\label{def:RMDP}
    A robust Markov decision process (RMDP) is a tuple $(S,\sinit,A,\cP,R)$, where the states $S$, initial state $\sinit$, actions $A$ and reward function $R$ are defined as in standard MDPs,
    and $\cP \colon \cU \to  (S \times A \parto \dist(S))$ is the uncertain transition function.
\end{definition}

Essentially, the uncertain transition function $\cP$ is a set of standard transition functions $P \colon S \times A \parto \dist(S)$, and we may thus also write $P \in \cP$ for a transition function $P$ that lies inside the uncertain transition function.

While strictly speaking not required, it is convenient to define the set of variables $X$ to have a unique variable for each possible transition of the RMDP, such that $X = \{x_{sas'} \mid (s,a,s') \in S \times A \times S\}$.
The uncertainty set $\cU$ is then a set of variable assignments, \ie, functions that map each variable to a real number, subject to constraints.
These constraints may, for example, define each variable's allowed range and encode dependencies between different variables.
Note that we do not explicitly add a constraint that each state-action pair is assigned a valid probability distribution but leave this implicit in the definition of the uncertain transition function $\cP$.
Alternatively, one can define RMDPs by having the uncertain transition function assign a function over the variables to each transition, effectively encoding the dependencies there, and having the uncertainty set only define the range of each variable. 
This construction would, however, require additional adjustments to move most of the discussion that follows (most notably around rectangularity) from the uncertainty set to the uncertain transition function.

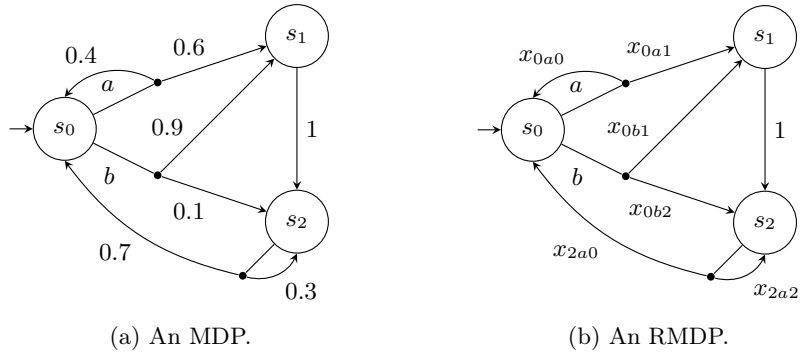
\begin{figure}[t]
\centering
    \begin{subfigure}[b]{0.49\textwidth}
    \centering
        \resizebox{0.8\linewidth}{!}{
            \begin{tikzpicture}[state/.append style={shape = ellipse}, >=stealth,
    bobbel/.style={minimum size=1mm,inner sep=0pt,fill=black,circle},
    mynode/.style={rectangle,fill=white,anchor=center}]]
    \node[state] (s0) at (1,0) {$s_0$};
    \node[state] (s1) at ($(s0) + (3,1.2)$) {$s_1$};
    \node[state] (s2) at ($(s0) + (3,-1.2)$) {$s_2$};
    \node[bobbel] (s0ba) at ($(s0) + (1.2,0.6)$) {};
    \node[bobbel] (s0bb) at ($(s0) + (1.2,-0.6)$) {};
    \node[bobbel] (s2b) at ($(s2) + (-0.7,-0.7)$) {};
    \draw[<-] (s0.west) -- +(-0.3,0);
    \draw (s0) -- node[above left]{$a$} (s0ba);
    \draw (s0) -- node[below left]{$b$} (s0bb);
    \draw (s0ba) edge[->] node[above left]{$0.6$} (s1);
    \draw (s0ba) edge[->, bend right = 40] node[above left]{$0.4$} (s0.north);
    \draw (s0bb) edge[->, pos = 0.45] node[above left, xshift=-2.5mm, yshift=-3mm]{$0.9$} (s1);
    \draw (s0bb) edge[->] node[below left]{$0.1$} (s2);
    \draw (s1) edge[->] node[right] {$1$} (s2);
    \draw (s2) -- node[below left]{} (s2b);
    \draw (s2b) edge[->, bend right = 40] node[below right]{$0.3$} (s2.south);
    \draw (s2b) edge[->, bend left = 20] node[below left]{$0.7$} (s0.south);
    \draw[white, line width = 0.2pt] ($(s0.west) +(-0.3,-2.3)$) -- (4.6,-2.3);
    \draw[white, line width = 0.2pt] (4.6,1.7) -- (4.6,-2.3);
\end{tikzpicture}
            }
        \caption{An MDP.}
        \label{fig:example:mdp}
    \end{subfigure}%
    \hfill
    \begin{subfigure}[b]{0.49\textwidth}
    \centering
        \resizebox{0.8\linewidth}{!}{
            \begin{tikzpicture}[state/.append style={shape = ellipse}, >=stealth,
    bobbel/.style={minimum size=1mm,inner sep=0pt,fill=black,circle},
    mynode/.style={rectangle,fill=white,anchor=center}]]
    \node[state] (s0) at (1,0) {$s_0$};
    \node[state] (s1) at ($(s0) + (3,1.2)$) {$s_1$};
    \node[state] (s2) at ($(s0) + (3,-1.2)$) {$s_2$};
    \node[bobbel] (s0ba) at ($(s0) + (1.2,0.6)$) {};
    \node[bobbel] (s0bb) at ($(s0) + (1.2,-0.6)$) {};
    \node[bobbel] (s2b) at ($(s2) + (-0.7,-0.7)$) {};
    \draw[<-] (s0.west) -- +(-0.3,0);
    \draw (s0) -- node[above left]{$a$} (s0ba);
    \draw (s0) -- node[below left]{$b$} (s0bb);
    \draw (s0ba) edge[->] node[above left]{$x_{0a1}$} (s1);
    \draw (s0ba) edge[->, bend right = 40] node[above left]{$x_{0a0}$} (s0.north);
    \draw (s0bb) edge[->, pos = 0.45] node[above left, xshift=-2.5mm, yshift=-3mm]{$x_{0b1}$} (s1);
    \draw (s0bb) edge[->] node[below left]{$x_{0b2}$} (s2);
    \draw (s1) edge[->] node[right] {$1$} (s2);
    \draw (s2) -- node[below left]{} (s2b);
    \draw (s2b) edge[->, bend right = 40] node[below right]{$x_{2a2}$} (s2.south);
    \draw (s2b) edge[->, bend left = 20] node[below left]{$x_{2a0}$} (s0.south);
    \draw[white, line width = 0.2pt] ($(s0.west) +(-0.3,-2.3)$) -- (4.6,-2.3);
    \draw[white, line width = 0.2pt] (4.6,1.7) -- (4.6,-2.3);
\end{tikzpicture}
        }
        \caption{An RMDP.}
        \label{fig:example_uncertainty_sets}
    \end{subfigure}
    \caption{An MDP and RMDP for \Cref{ex:uncertainty:sets}.}
    \label{fig:example_mdp_rmdp}
\end{figure}

\begin{example}\label{ex:uncertainty:sets}
    \Cref{fig:example_mdp_rmdp} depicts an MDP and an RMDP.
    Below are three possible uncertainty sets for this RMDP: 
    \begin{align}
        &    \cU_1 = \left\{{x_{0a1}} \in [0.1,0.9] \wedge {x_{0b1}} \in [0.1,0.9] \wedge {x_{2a0}} \in [0.1,0.9]\right\},\notag \\
        & \cU_2 = \left\{{x_{0a1}} \in [0.1,0.4] \wedge {x_{0b1} = 2x_{0a1}} \wedge {x_{2a0}} \in [0.1,0.9]\right\}, \label{eq:ex1:uncertainty:sets}\\
        & \cU_3 = \left\{{x_{0a1}} \in [0.1,0.4] \wedge {x_{0b1} = 2x_{0a1}} \wedge {x_{2a0} = x_{0a1}}\right\}. \notag
    \end{align}
    The agent can choose between action $a$ and $b$ in state $s_0$ and has singleton choices in the other states.
    An adversary, \emph{nature}, chooses variable assignments for $x_{0a0}, x_{0a1}, x_{0b1}, x_{0b2}$, $x_{2a0}$, and $x_{2a2}$.
    As mentioned above, the restriction that each state-action pair is assigned a valid probability distribution is implied by the definition of the uncertain transition function $\cP$.
    We can therefore focus purely on the choices of $x_{0a0}, x_{0b1}$, and $x_{2a0}$.
    
    The uncertainty sets give restrictions on the possible variable assignments.
    In uncertainty set $\cU_1$, variables $x_{0a0}, x_{0b1}$, and $x_{2a0}$ can each be given any value in the interval $[0.1,0.9]$.
    A possible variable assignment in $\cU_1$ is $f^1 = \{x_{0a0} \mapsto 0.3, x_{0b1} \mapsto 0.1, x_{2a0} \mapsto 0.8\}$.
    This variable assignment is not possible in uncertainty sets $\cU_2$ and $\cU_3$ because of the dependencies between the variables.
    For example, in uncertainty set $\cU_2$, variable $x_{0b1}$ must be assigned twice the value of $x_{0a0}$, whose value must now be in the interval $[0.1,0.4]$.
    A possible variable assignment in $\cU_2$ is $f^2 = \{x_{0a0} \mapsto 0.3, x_{0b1} \mapsto 0.6, x_{2a0} \mapsto 0.8\}$.
    We further discuss the effect of dependencies in the uncertainty set in \Cref{subsec:rmdp:semantics}.
\end{example}

\subsubsection*{Objectives.}
For ease of presentation, we again focus on reach-reward maximization as an objective.
However, as there is no single transition function, the goal is now to compute an optimal \emph{robust} policy, meaning optimal against the worst-case probabilities in the model. 
What this worst-case exactly is, depends on the semantics of the RMDP.

\subsection{RMDP Semantics and Structural Assumptions}\label{subsec:rmdp:semantics}

RMDPs can be seen as a game between the \emph{agent}, who aims to maximize their reward by selecting actions, and an adversarial \emph{nature}, who aims to minimize the agent's reward by selecting variable assignments from the uncertainty set.
Nature hence simulates the worst-case transition function that the agent should be robust against.
This game interpretation can be fully formalized into a zero-sum stochastic game (SG), as we shall discuss further in \Cref{subsec:stochastic:games}.

Intuitively, the game is constructed by adding a new set of states $S \times A$ for nature that consists of tuples of the state-action pairs the agent was in.
At each such state-action pair, nature selects a variable assignment from the uncertainty set that determines the transition function $P \in \cP$.

The precise rules of the game, and with that the semantics of RMDPs, that determine which variable assignments nature is allowed to choose are controlled by two factors: (1) possible dependencies between nature's choice of the variable assignments between different states or actions, known as (non)-rectangularity; and (2) whether previous choices by nature restrict its future choices, known as the static and dynamic uncertainty semantics.
These two factors determine the available policies, \ie, transition functions, for nature, and thus the worst-case transition function that the agent must be robust against.
We now discuss both concerns in more detail.

Dependencies between the variables, or lack thereof, immediately follow from the constraints used to define the uncertainty set $\cU$.
\emph{Independence} between states or state-action pairs is commonly referred to as \emph{rectangularity}.
Informally, an uncertainty set $\cU$ is state-action or $(s,a)$-rectangular if there are no dependencies between the constraints on the variables at different state-action pairs, and state or $s$-rectangular if there are no dependencies between constraints on the variables at different states.
More formally, following standard notation~\cite{DBLP:journals/mor/WiesemannKR13}:
\begin{definition}[Rectangularity]
    The uncertainty set $\cU$ is \emph{$(s,a)$-rectangular} if it can be split into lower dimensional uncertainty sets $\cU_{(s,a)}$ that only relate to the variables at the respective state-action pair $(s,a)$, such that their product forms the whole uncertainty set: $\cU = \bigtimes_{(s,a) \in S \times A} \cU_{(s,a)}$.
    Similarly, an uncertainty set $\cU$ is \emph{$s$-rectangular} if $\cU$ can be split into lower dimensional uncertainty sets $\cU_{(s)}$ that only relate to variables at state $s$, such that $\cU = \bigtimes_{s \in S} \cU_{(s)}$.
\end{definition}

\begin{example}
    We revisit the RMDP in \Cref{fig:example_uncertainty_sets} and the three possible uncertainty sets in \Cref{eq:ex1:uncertainty:sets}.
    The set $\cU_1$ is an $(s,a)$-rectangular uncertainty set, as each variable influences the transition probabilities in only one state-action pair.
    In other words, there are no dependencies between constraints on the variables at different state-action pairs.
    In $\cU_2$ the transition probabilities for state-action pairs $\tup{s_0,a}$ and $\tup{s_0,b}$ both depend on variable $x_{0a1}$.
    Therefore, $\cU_2$ no longer has independence between actions but is still $s$-rectangular.
    The final uncertainty set, $\cU_3$, has dependencies between all variables and is, therefore, \emph{non-rectangular}.  
\end{example}

The type of rectangularity has, together with whether the uncertainty set is convex or not, direct consequences for the computational complexity of policy evaluation, \ie, computing the value for a given policy, and the type of policy that is sufficient to be optimal for discounted reward objectives under static uncertainty semantics.
These results are due to Wiesemann et al.~\cite{DBLP:journals/mor/WiesemannKR13} and presented in \Cref{tab:rmdp:policy:complexity}.

Under $s$-rectangularity, an additional assumption is made that nature can no longer observe the last action of the agent.
This assumption is mentioned explicitly in~\cite{DBLP:conf/icml/HoPW18,DBLP:journals/jmlr/HoPW21,DBLP:journals/mor/WiesemannKR13} but often left implicit.
Note that this assumption on the last-action observability does influence the optimal policy and value, as demonstrated by \Cref{ex_s-rec_act_obs}. 
The question of last-action observability corresponds to the difference between \emph{agent first} and \emph{nature first} semantics in~\cite{bovy2024imprecise}.

\begin{figure}[t]
    \centering
    \resizebox{0.6\textwidth}{!}{
    \begin{tikzpicture}[state/.append style={shape = circle}, >=stealth,
    bobbel/.style={minimum size=1mm,inner sep=0pt,fill=black,circle}]
    \node[state] (s0) at (1,0) {$s_0$};
    \node[state] (s1) at ($(s0) + (2.5,1.2)$) {$s_1$};
    \node[state] (s2) at ($(s0) + (2.5,-1.2)$) {$s_2$};
    \node[state] (s3) at ($(s1) + (2.5,0)$) {$s_3$};
    \node[state] (s4) at ($(s2) + (2.5,0)$) {$s_4$};
    \node[bobbel] (s0ba) at ($(s0) + (0.75,0.6)$) {};
    \node[bobbel] (s0bb) at ($(s0) + (0.75,-0.6)$) {};
    \draw[<-] (s0.west) -- +(-0.3,0);
    \draw (s0) edge[-] node[above]{$a$} (s0ba);
    \draw (s0) edge[-] node[below]{$b$} (s0bb);
    \draw (s0ba) edge[->] node[above left] {$x_{0a1}$} (s1);
    \draw (s0ba) edge[->, pos = 0.7] node[above right] {$x_{0a2}$} (s2);
    \draw (s0bb) edge[->, pos = 0.7] node[below right] {$x_{0b1}$} (s1);
    \draw (s0bb) edge[->, pos = 0.5] node[below left] {$x_{0b2}$} (s2);
    \draw (s1) edge[->] node[above] {$R = 50$} (s3);
    \draw (s2) edge[->] node[above] {$R = 100$} (s4);
    \draw (s3) edge[loop right] node[right]{$1$} (s4);
    \draw (s4) edge[loop right] node[right]{$1$}(s3);
\end{tikzpicture}
    }
    \caption{An RMDP for \Cref{ex_s-rec_act_obs}.}
    \label{fig:fig_s-rec_act_obs}
\end{figure}
\begin{example}[Last-action observability]\label{ex_s-rec_act_obs}
    \Cref{fig:fig_s-rec_act_obs} depicts an RMDP.
    Below is an $s$-rectangular uncertainty set:
    \begin{align*}
        & \cU = \left\{x_{0a1} \in [0.1,0.9] \wedge x_{0a1} = x_{0b2}\right\}.
    \end{align*}
    Whether or not nature observed the agent's last action determines whether or not nature has to take the dependency between $x_{0a1}$ and $x_{0b2}$ into account.
    If nature observes the agent's last action, it can achieve an expected reward of $55$ by choosing the maximal $x_{0a1} = x_{0b2} = 0.9$ when observing action $a$, and by choosing the minimal $x_{0a1} = x_{0b2} = 0.1$ when observing action $b$. 
    If nature does not have this information, it has to account for both possible agent actions.
    The best course of action for nature is then to choose $x_{0a1} = x_{0b2} = 0.5$, leading to an expected reward of $75$ regardless of the agent's choice.
\end{example}

\subsubsection{Static and dynamic uncertainty semantics. }
The second point about RMDP semantics is whether nature's previous choice at a certain state-action pair should restrict its possible future choices.
To that end, Iyengar~\cite{DBLP:journals/mor/Iyengar05} introduced the notions of \emph{static} and \emph{dynamic} uncertainty semantics.
Static uncertainty semantics require nature to play a `once-for-all' policy: if the state-action pair is revisited, nature is required to use the same variable assignment from the uncertainty set as before.
In contrast, under dynamic uncertainty semantics nature plays `memoryless' and is free to choose any variable assignment at every step.
Simultaneous but independently, Nilim and El Ghaoui introduced these semantics as \emph{time-stationary} and \emph{time-varying} uncertainty models~\cite{DBLP:journals/ior/NilimG05}.
Note that these notions have only been introduced for $(s,a)$-rectangular RMDPs and are only of concern in cyclic, infinite horizon models.
Interestingly, Iyengar~\cite{DBLP:journals/mor/Iyengar05} also shows that the distinction between static and dynamic uncertainty does not matter for reward maximization in $(s,a)$-rectangular RMDPs.
A similar result was established for reachability in interval Markov chains in~\cite{DBLP:journals/ipl/ChenHK13}.
We state the result in general in the following lemma.

\begin{lemma}[Static and dynamic uncertainty coincide~\cite{DBLP:journals/mor/Iyengar05}]
    Consider an $(s,a)$-rectangular RMDP where both agent and nature are restricted to stationary policies, \ie, policies of type $\policy \colon S \to \dist(A)$.
    Let $\policy^\text{st}$ be the optimal policy under static uncertainty, and $\policy^\text{dy}$ be the optimal policy under dynamic uncertainty semantics.
    The robust values of these policies coincide, \ie, $V_{\policy^\text{st}} = V_{\policy^\text{dy}}$.
\end{lemma}

\begin{table}[t]
    \centering
    \begin{tabular}{@{\hskip 0.1cm}l@{\hskip 0.4cm}l@{\hskip 1.2cm}l@{\hskip 0.6cm}l@{\hskip 0.1cm}}
    \toprule
        \multicolumn{2}{l}{Uncertainty set \& rectangularity}  & Optimal policy class & Complexity \\
         \midrule
         & $(s,a)$-rectangular & Stationary, deterministic & Polynomial\\
        Convex & $s$-rectangular & Stationary, randomized & Polynomial \\
         & non-rectangular & History, randomized & NP-hard \\
             \midrule
         & $(s,a)$-rectangular & Stationary, deterministic & NP-hard\\
        Nonconvex & $s$-rectangular & History, randomized & NP-hard\\
         & non-rectangular & History, randomized & NP-hard\\
         \bottomrule \\
    \end{tabular}
         \caption{Policy classes that are sufficient and computational complexity of policy evaluation for discounted reward RMDPs with various types of uncertainty sets, as identified by Wiesemann et al.~\cite{DBLP:journals/mor/WiesemannKR13} assuming \emph{static} uncertainty semantics.}
    \label{tab:rmdp:policy:complexity}
\end{table}

\subsection{Robust Dynamic Programming}\label{subsec:robustdp}

In this section, we discuss how value iteration and policy iteration can be adapted rather straightforwardly for $(s,a)$-rectangular RMDPs.

\begin{remark}[Graph preservation]\label{remark:graph:pres}
    For computational tractability of robust dynamic programming, especially of objectives that rely on preprocessing the underlying graph, such as the reach-reward objective we consider, it is often assumed that the uncertainty set should be \emph{graph preserving}. 
    That is, all variable assignments in the uncertainty set $\cU$ imply the same topology for the underlying graphs.
    Hence, if there exists some $P \in \cP$ with $P(s,a,s') = 0$ for some transition, then all other $P' \in \cP$ should also have $P'(s,a,s') = 0$. 
\end{remark}

Recall \Cref{eq:bellman:mdp}, describing value iteration in a standard MDP.
In an RMDP, we do not have access to a precisely defined transition function $P \colon S \times A \parto \dist(S)$.
Instead, we have the uncertain transition function $\cP$ that defines a set of such transition functions $P \in \cP$.

\emph{Robust value iteration} adapts value iteration by accounting for the worst-case $P \in \cP$ at each iteration.
This is achieved by replacing the inner sum $\sum_{s' \in S} P(s,a,s')V^n(s')$ by an \emph{inner minimization problem}:
\begin{align}
    \Vpes^{(n+1)}(s) =  \max_{a \in A} \left\{ R(s,a) +  \inf_{P \in \cP} \left\{  \sum_{s' \in S} P(s,a,s')\Vpes^{(n)}(s') \right\} \right\}.\label{eq:rdp:disc}
\end{align}
We write $\Vpes$ instead of $V$, which is now the \emph{worst-case} or \emph{pessimistic} value of the RMDP.
That is, $\Vpes$ is a lower bound on the value the agent can possibly achieve.
\emph{Best-case} or \emph{optimistic} interpretations also exist, which we discuss later.

Under our assumption that the uncertainty set $\cU$ is $(s,a)$-rectangular, we may replace the global minimization problem $\inf_{P \in \cP}$ by a local one: 
\begin{align}
    \Vpes^{(n+1)}(s) =  \max_{a \in A} \left\{ R(s,a) +  \inf_{P(s,a) \in \cP(s,a)} \left\{  \sum_{s' \in S} P(s,a,s')\Vpes^{(n)}(s') \right\} \right\}. \label{eq:rdp:sa:disc}    
\end{align}
If, additionally, the uncertainty set $\cU$ is convex, for instance, because all constraints are linear, the inner minimization problem can be solved efficiently via, \eg, convex optimization methods.
Hence, robust value iteration extends regular value iteration by solving an additional inner problem at every iteration.
In general, the computational tractability of RMDPs primarily relies on whether or not this inner problem is efficiently solvable.

As discussed in \Cref{subsec:rmdp:semantics} and shown in \Cref{tab:rmdp:policy:complexity}, stationary deterministic policies are sufficient for optimality in $(s,a)$-rectangular RMDPs with convex uncertainty sets.
Thus, an optimal \emph{robust policy} $\policypes^*$ can again be extracted via 
\begin{align}
\policypes^*(s) = \argmax_{a \in A} \left\{ R(s,a) +  \inf_{P(s,a) \in \cP(s,a)} \left\{ \sum_{s' \in S} P(s,a,s') \Vpes^*(s') \right\} \right\}.\label{eq:rdp:policy:disc}
\end{align}
We underline the policy $\policypes^*$ to denote that it is an optimal \emph{robust} policy, as we later also touch upon optimal \emph{optimistic} policies, which we shall denote by $\policyopt^*$.

\emph{Robust policy iteration}~\cite{DBLP:journals/mor/Iyengar05,DBLP:journals/informs/KaufmanS13} extends standard policy iteration in a similar way.
Policy evaluation, \ie, model checking the induced \emph{robust} Markov chain, is done by performing robust dynamic programming for some stationary policy $\policy$:
\[
\Vpes^{(n+1)}_{\policy} = \sum_{a \in A} \policy(s,a) \cdot \left( R(s,a) + \inf_{P(s,a) \in \cP(s,a)} \left\{ \sum_{s' \in S} P(s,a,s')\Vpes^{(n)}_{\policy}(s') \right\}  \right). 
\]
Here, we again use that our uncertainty set is $(s,a)$-rectangular and convex to ensure an efficiently solvable inner minimization problem.
After convergence, we use the robust state values under the current policy $\Vpes_{\policy}^*$ to compute the robust state-action values $\Qpes_{\policy}$:
\[
\Qpes_{\policy}(s,a) = R(s,a) +  \inf_{P(s,a) \in \cP(s,a)} \left\{ \sum_{s' \in S} P(s,a,s') \Vpes_{\policy}^*(s') \right\}.
\]
The policy improvement step is performed on these robust state-action values: 
\[
\policy'(s) = \argmax_{a \in A} \Qpes_{\policy}(s,a).
\]
The process repeats until the policy stabilizes, \ie, $\policy' = \policy$, after which an optimal robust policy $\policypes^* = \policy'$ has been found.

\subsubsection{Optimistic dynamic programming.}
Instead of assuming the worst-case from the uncertainty set, we may also assume the agent and nature play cooperatively. 
That is, both players attempt to maximize the agent's reward, and we instead obtain \emph{optimistic} values $\Vopt$ that are computed in the same way as the pessimistic values were, except that the inner minimization problem from \Cref{eq:rdp:disc} is now replaced by an inner \emph{maximization} problem:
\[
    \Vopt^{(n+1)}(s) =  \max_{a \in A} \left\{ R(s,a) + \sup_{P \in \cP} \left\{  \sum_{s' \in S} P(s,a,s')\Vopt^{(n)}(s') \right\} \right\}.%
\]
The optimal \emph{optimistic} policy $\policyopt^*$ is again extracted by one final step of dynamic programming, as in \Cref{eq:rdp:policy:disc}:
\[
\policyopt^*(s) = \argmax_{a \in A} \left\{ R(s,a) + \sup_{P \in \cP} \left\{ \sum_{s' \in S} P(s,a,s') \Vopt^*(s')  \right\} \right\}.
\]

\subsubsection{Convex optimization.}
Since dynamic programming approaches for MDPs extend with relative ease to RMDPs, especially in the case of $(s,a)$-rectangular uncertainty sets, a natural question to ask is whether the same goes for convex optimization approaches, and in particular the linear programming (LP) formulation for MDPs.
As Iyengar~\cite{DBLP:journals/mor/Iyengar05} already notes, however, that is not the case, and the natural analogue of the LP of MDPs for RMDPs yields, in fact, a nonconvex optimization problem.
In contrast, the optimistic setting does yield tractable LPs via standard dualization techniques, which have been applied to solve PCTL objectives in $(s,a)$-rectangular RMDPs with convex uncertainty sets~\cite{DBLP:conf/cav/PuggelliLSS13}.

\subsubsection{Methods for $s$-rectangular RMDPs.}
For $s$-rectangular RMDPs, dynamic programming does not extend so straightforwardly, and a lot of research has focused on finding efficient Bellman operators for various types of uncertainty sets.
Most notably, $s$-rectangular \Lone-MDPs~\cite{DBLP:conf/icml/HoPW18}, but also $s$-rectangular uncertainty sets defined by an $L_{\infty}$-norm~\cite{DBLP:conf/nips/BehzadianPH21} or $\phi$-divergences~\cite{DBLP:conf/nips/HoPW22}.
In~\cite{DBLP:journals/jmlr/HoPW21}, a policy iteration algorithm was introduced, while~\cite{DBLP:conf/aaai/GadotDKELM24,DBLP:conf/nips/KumarDGLM23,DBLP:conf/icml/WangHP23} employ policy gradient techniques.

\subsubsection{Other objectives.} 
The RMDP literature primarily focuses on either finite horizon or discounted infinite horizon reward maximization.
Adaptation to reachability objectives, such as the reach-reward maximization we consider, is usually straightforward, provided the graph preservation property of \Cref{remark:graph:pres} is met.
Temporal logic objectives can be reduced to such reach-reward objectives via a product construction~\cite{DBLP:conf/cdc/WolffTM12}.
Finally, recent works study average reward (also known as mean payoff) and Blackwell optimality in RMDPs~\cite{DBLP:journals/corr/abs-2312-13912,DBLP:conf/nips/Grand-ClementP23,DBLP:journals/corr/abs-2312-03618}. 
Average reward objectives consider the problem of maximizing the average reward collected in $t$ time steps when $\lim_{t \to \infty}$, and Blackwell optimality balances the standard discounted reward objective by also accounting for long-term reward.
A policy is Blackwell optimal if it is optimal for all discount factors sufficiently close to one, \ie, all $\gamma \in [\gamma^*, 1)$~\cite{DBLP:books/wi/Puterman94}.

\subsection{Well-Known RMDP Instances}\label{subsec:rmdp:instances}
We review common types of RMDPs often used in formal verification and AI, namely interval MDPs, \Lone-MDPs, and multi-environment MDPs.
We provide a more commonly used tuple definition for each and explain how it fits the general RMDP framework.

\subsection*{Interval MDPs}
Interval MDPs (IMDPs)~\cite{DBLP:journals/ior/NilimG05}, also referred to as bounded-parameter MDPs~\cite{DBLP:journals/ai/GivanLD00} or uncertain MDPs~\cite{DBLP:conf/nips/SuilenS0022,DBLP:conf/cdc/WolffTM12}, are a special instance of $(s,a)$-rectangular RMDPs.

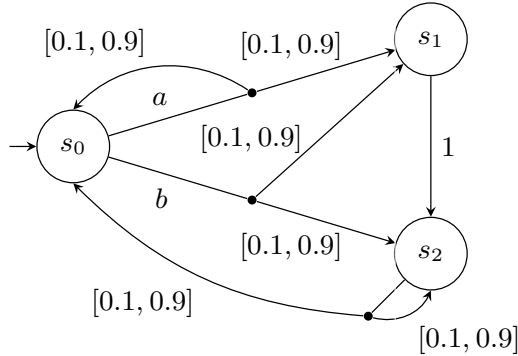
\begin{figure}[t]
    \centering
    \resizebox{0.6\textwidth}{!}{
    \begin{tikzpicture}[state/.append style={shape = ellipse}, >=stealth,
    bobbel/.style={minimum size=1mm,inner sep=0pt,fill=black,circle},
    mynode/.style={rectangle,fill=white,anchor=center}]]
    \node[state] (s0) at (1,0) {$s_0$};
    \node[state] (s1) at ($(s0) + (4,1.2)$) {$s_1$};
    \node[state] (s2) at ($(s0) + (4,-1.2)$) {$s_2$};
    \node[bobbel] (s0ba) at ($(s0) + (2,0.6)$) {};
    \node[bobbel] (s0bb) at ($(s0) + (2,-0.6)$) {};
    \node[bobbel] (s2b) at ($(s2) + (-0.7,-0.7)$) {};
    \draw[<-] (s0.west) -- +(-0.3,0);
    \draw (s0) -- node[above left]{$a$} (s0ba);
    \draw (s0) -- node[below left]{$b$} (s0bb);
    \draw (s0ba) edge[->] node[above left, xshift=3mm]{$[0.1,0.9]$} (s1);
    \draw (s0ba) edge[->, bend right = 40] node[above left]{$[0.1,0.9]$} (s0.north);
    \draw (s0bb) edge[->, pos = 0.45] node[above left, xshift=-1mm, yshift=-3mm]{$[0.1,0.9]$} (s1);
    \draw (s0bb) edge[->] node[below left, xshift=3mm]{$[0.1,0.9]$} (s2);
    \draw (s1) edge[->] node[right] {$1$} (s2);
    \draw (s2) -- node[below left]{} (s2b);
    \draw (s2b) edge[->, bend right = 40] node[below right]{$[0.1,0.9]$} (s2.south);
    \draw (s2b) edge[->, bend left = 20] node[below left]{$[0.1,0.9]$} (s0.south);
    \draw[white, line width = 0.2pt] ($(s0.west) +(-0.3,-2.3)$) -- (4.6,-2.3);
    \draw[white, line width = 0.2pt] (4.6,1.7) -- (4.6,-2.3);
\end{tikzpicture}
    }
    \caption{An example IMDP.}
    \label{fig:ex:IMDP}
\end{figure}

\begin{definition}[IMDP]
    An interval MDP (IMDP) is a tuple $(S,\sinit,A,\Plow,\Pup, R)$, where $\check{P} \colon S \times A \times S \parto [0,1]$ and $\hat{P} \colon S \times A \times S \parto [0,1]$ are two transition functions that assign lower and upper bounds to each transition, respectively, such that $\check{P} \leq \hat{P}$ and for all transitions $\check{P}(s,a,s') = 0 \iff \hat{P}(s,a,s') = 0$.
\end{definition} 
Our definition of an IMDP requires that a transition either does not exist (where both $\check{P}$ and $\hat{P}$ are zero) or is assigned an interval with a non-zero lower bound, thus ensuring graph preservation for tractable (standard) robust dynamic programming (\Cref{remark:graph:pres}).
For IMDPs, however, the statistical model checking literature offers solutions to circumvent the need for this requirement~\cite{DBLP:conf/cav/AshokKW19,DBLP:conf/tacas/DacaHKP16}. 

IMDPs have a constraint that each state-action pair is required to have a valid probability distribution.
An IMDP is an RMDP $(S, \sinit,A,\cP,R)$ where the uncertainty set is of the form $\cU = \{f \colon X \to \real \mid \forall (s,a,s') \in S \times A \times S,~ f(x_{sas'}) \in [i,j]_{sas'} \subseteq [0,1] \wedge \forall (s,a) \in S \times A,~\sum_{s' \in S} f(x_{sas'}) = 1 \}$.
An example IMDP is depicted in \Cref{fig:ex:IMDP}.
Note that this IMDP is precisely the RMDP from \Cref{fig:example_mdp_rmdp} with uncertainty set $\cU^1$, see \Cref{ex:uncertainty:sets}.

IMDPs have the nice property that their inner problem can be solved efficiently via a bisection algorithm~\cite{DBLP:journals/ior/NilimG05}, more explicitly given for interval DTMCs in~\cite{DBLP:journals/jlp/KatoenKLW12} and presented in \Cref{alg:imdp:algorithm}.
This algorithm sorts the successor states $S' = \{s_1',\dots,s_m'\}$ of state-action pair $(s,a)$ by the current value $V^{(n)}$ in ascending order.
A variable $\mathit{budget}$ indicates how much probability mass is still free to assign when we start with assigning the lower bounds to each successor state. 
Successor states occurring at low indices, \ie, with low values $V^{(n)}$, will be assigned the upper bound of the transition leading to them until the budget runs out.
One state will get a remaining $\mathit{budget}$ added to its lower bound, which is the first state for which it is no longer possible to replace the lower bound by the upper bound, ensuring the transition probability lies within its interval. 
The remaining successor states, with high values, will be assigned the lower bounds, as the budget for replacement is now zero.
As a result, transition function $P(s,a,\cdot)$ forms a valid probability distribution.

For optimistic dynamic programming, \ie, the case where the inner problem is given by the supremum over the uncertainty set instead of the infimum, we only need to reverse the order in which the states are sorted in line 1 of \Cref{alg:imdp:algorithm}.

\begin{algorithm}[t]
    \caption{Algorithm to solve the IMDP inner problem $\inf_{P \in \cP(s,a)}$}\label{alg:imdp:algorithm}
    \begin{algorithmic}[1]
        \STATE  Sort $S' = \{s'_1,\dots, s'_m\}$ according to $V^{(n)}$ ascending such that $V^{(n)}(s'_i) \leq V^{(n)}(s'_{i+1})$
        \STATE $\forall s_i' \in S'$: $P(s,a,s'_i) \gets 0$
        \STATE $\mathit{budget} = 1 - \sum_{s' \in S'} \Plow(s,a,s')$
        \STATE $i \gets 1$ 
        \WHILE {$\mathit{budget}  - (\Pup(s,a,s'_i) - \Plow(s,a,s'_i)) \geq 0$} 
            \STATE $P(s,a,s'_i) \gets \Pup(s,a,s'_i)$  
            \STATE $\mathit{budget}  \gets \mathit{budget}  - (\Pup(s,a,s'_i) - \Plow(s,a,s'_i))$
            \STATE $i \gets i+1$
        \ENDWHILE
        \STATE $P(s,a,s'_i) \gets  \Plow(s,a,s'_i) + \mathit{budget}$
        \STATE $\forall j \in \{i+1, \dots, m\}$: $P(s,a,s_j') \gets \Plow(s,a,s_j')$ 
        \RETURN $P(s,a,\cdot)$
    \end{algorithmic}
\end{algorithm}

\subsection*{\Lone-MDPs}
\Lone-MDPs~\cite{DBLP:journals/jcss/StrehlL08} are another instance of $(s,a)$-rectangular RMDPs.
Where IMDPs put an error margin around each individual transition probability, \Lone-MDPs put an error margin around each probability distribution at a state-action pair.

\begin{definition}[\Lone-MDP]
    An \Lone-MDP is a tuple $(S,\sinit, A, \Pest, R, d)$, where $S$, $\sinit \in S$, $A$ and $R$ are as for standard MDPs, $\Pest \colon S \times A \parto \dist(S)$ is the centre transition function and $d \colon S \times A \to \bR_{\geq 0}$ is a distance function assigning an error bound to each state-action pair.
\end{definition}
An \Lone-MDP is an RMDP $(S, \sinit,A,\cP,R)$ where the uncertainty set is given by $\cU = \{f \colon X \to \real \mid \forall (s,a) \in S \times A,~ \sum_{s' \in S} | f(x_{sas'}) - \tilde{P}(s,a,s') | \leq d(s,a) \}$, where $d(s,a)$ bounds the \Lone-error between the reference distribution $\Pest(s,a)$ and all other distributions $P(s,a) \in \cP(s,a)$.

Similar to IMDPs, the inner optimization problem for \Lone-MDPs can be solved efficiently, again by ordering the successor states along their current value and assigning low-ranking states the most possible probability mass and high-ranking states the least probability mass.
Specifically, the state with the lowest value $s'_s$ gets probability mass $\nicefrac{d(s,a)}{2}$ added to its estimate $\Pest(s,a,s'_1)$, and the remaining states from high to low get probability mass subtracted up to a total of $\nicefrac{d(s,a)}{2}$, ensuring a valid probability distribution.
This algorithm is the dual of the algorithm for computing the optimistic inner problem $\sup_{P \in \cP(s,a)}$ of~\cite{DBLP:journals/jcss/StrehlL08} and explicitly given in \Cref{alg:lone:algorithm}.

 \begin{algorithm}[t]
     \caption{Algorithm to solve the \Lone-MDP inner problem $\inf_{P \in \cP(s,a)}$}\label{alg:lone:algorithm}
     \begin{algorithmic}[1]
         \STATE  Sort $S' = \{s'_1,\dots, s'_m\}$ according to $V^{(n)}$ ascending such that $V^{(n)}(s'_i) \leq V^{(n)}(s'_{i+1})$
         \STATE $P(s,a,s_1') \gets \min\{1, \Pest(s,a,s_1') +  \nicefrac{d(s,a)}{2}\}$
         \STATE $\forall s'_i \neq s_1' \in S'$: $P(s,a,s'_i) \gets \Pest(s,a,s_i')$
         \STATE $i \gets m$
         \WHILE {$\sum_{j = 1}^m P(s,a,s_j') > 1$} 
             \STATE $P(s,a,s'_i) \gets \max\{ 0, 1 - \sum_{j \in \{1,\dots,m\} \setminus \{i\}} P(s,a,s_j') \} $  
             \STATE $i \gets i-1$
         \ENDWHILE
         \RETURN $P(s,a,\cdot)$
     \end{algorithmic}
 \end{algorithm}

\subsection*{Multi-Environment MDPs}
Multi-environment MDPs (MEMDPs)~\cite{DBLP:conf/fsttcs/RaskinS14} model \emph{discrete} uncertainty. 
Specifically, a MEMDP is a finite set of MDPs that share the same states, actions, and reward function and only differ in their transition functions.
Each MDP in a MEMDP is called an \emph{environment}.

\begin{definition}[MEMDP]
A multi-environment MDP (MEMDP) is a tuple $(S,\sinit, A, \{P_i\}_{i \in I}, R)$ where $S, \sinit, A, R$ are as for MDPs, and $\{P_i \colon S \times A \parto \dist(S)\}_{i \in I}$ is a set of $I = \{1,\dots,n\}$ transition functions that are consistent with each other in terms of enabled actions: $\forall (s,a) \in S \times A, \forall i,j \in I, \; P_i(s,a) = \bot \iff P_j(s,a) = \bot$. 
\end{definition}
A MEMDP is an RMDP $(S,\sinit,A,\cP,R)$ where the uncertainty set $\cU$ is discrete: $|\cU| \neq \infty$, and $\cP = \{ P_i \}_{i \in I}$.
In general, MEMDPs are \emph{non-rectangular} and follow \emph{static} uncertainty semantics, as nature's choices at each state-action pair must be consistent with, and equivalent to, choosing a single $P_i \in \cP$ at the start.
As the uncertainty set is discrete, it is also nonconvex. 
Hence, optimal policies in MEMDPs need to be history-based and randomized; see \Cref{tab:rmdp:policy:complexity}.

MEMDPs have been studied in both formal methods and AI.
In AI, MEMDPs have caught interest because of their applications in robotics, naturally modelling several possible worlds a robot may act in~\cite{DBLP:conf/aaai/RigterLH21}.
Besides being RMDPs, MEMDPs are also a subclass of \emph{partially observable} MDPs (POMDPs), where the agent does not directly observe the states~\cite{DBLP:journals/ai/KaelblingLC98}.
In particular, a MEMDP can be transformed into a POMDP by taking the disjoint union of all environments and using the partial observability to hide in which environment the agent is playing~\cite{DBLP:conf/aips/ChatterjeeCK0R20}.
As a result, quantitative objectives such as reward maximization may be solved by casting the MEMDP to a POMDP and using off-the-shelf POMDP methods.

In formal methods, emphasis has been given to complexity results, especially for \emph{almost-sure} objectives, \ie, objectives that need to be satisfied with probability one.
In~\cite{DBLP:conf/fsttcs/RaskinS14}, it is shown that almost-sure parity objectives are in $\sP$ for MEMDPs of two environments, while~\cite{DBLP:conf/tacas/VegtJJ23} shows that already for almost-sure reachability, an arbitrary number of environments leads to $\sPSPACE$-completeness.
Recent work completes the complexity landscape for qualitative objectives in MEMDPs by establishing $\sPSPACE$-completeness for almost-sure parity and Rabin objectives~\cite{suilen2024pspace}.
In contrast, almost-sure reachability is already $\sEXP$-complete for POMDPs~\cite{DBLP:conf/mfcs/ChatterjeeDH10}, and almost-sure parity or Rabin objectives are undecidable~\cite{DBLP:conf/fossacs/BaierBG08},
showing that MEMDPs are an interesting class worth investigating.
\section{Connections to Other Models}\label{sec:connections}
In the following, we summarize the connections between RMDPs and some other commonly used models in formal methods and AI.
In particular, we highlight the connections with parametric MDPs (\Cref{subsec:pmdp}), stochastic games (\Cref{subsec:stochastic:games}), robust POMDPs (\Cref{subsec:rpomdp}), and a range of models that assume additional distributional information over the parameters (\Cref{subsec:likelihoods}).

\subsection{Parametric MDPs}\label{subsec:pmdp}

Our general definition of RMDPs as given in \Cref{def:RMDP} closely resembles that of \emph{parametric} MDPs (pMDPs)~\cite{DBLP:conf/birthday/0001JK22}.
Indeed, both models assign variables (parameters) to the transitions instead of concrete probabilities, effectively defining a set of possible MDP models.
Typically, pMDPs are defined more generally and allow for a rational function over two polynomials on the transitions, encoding dependencies between transitions directly by having two of these rationals share some of the parameters.

The \emph{parameter synthesis} problem~\cite{DBLP:conf/mbmv/DehnertJ0CVKAB16} typically considered in pMDPs is, however, different from the problem of computing robust policies and values in RMDPs.
Parameter synthesis asks whether there exists a variable assignment such that \emph{for all} policies, a certain (reachability) specification is met.
That is, the quantifiers are reversed compared to RMDPs.

Common techniques for parameter synthesis in pMDPs or pMCs include convex optimization approaches~\cite{DBLP:conf/tacas/Cubuktepe0JKPPT17,DBLP:conf/atva/CubuktepeJJKT18,DBLP:journals/tac/CubuktepeJJKT22}, parameter lifting~\cite{DBLP:conf/atva/QuatmannD0JK16}, or exact computation of the solution function~\cite{DBLP:journals/fmsd/JungesAHJKQV24}.
Tool support for pMDPs can be found in, \eg, \Storm~\cite{DBLP:journals/sttt/HenselJKQV22} or \Prophesy~\cite{DBLP:conf/cav/DehnertJJCVBKA15}.
Parameter synthesis in pMDPs with memoryless deterministic policies is known to be $\sNP$-complete for a fixed number of parameters and $\sETR$-complete for arbitrary numbers of parameters~\cite{DBLP:journals/jcss/JungesK0W21}.

\subsection{Stochastic Games}\label{subsec:stochastic:games}
The connection between \emph{stochastic games} (SG) and RMDPs has been noted many times in the RMDP literature. 
See, \eg,~\cite{DBLP:journals/mor/GoyalG23,DBLP:journals/corr/abs-2312-03618,DBLP:journals/mor/Iyengar05,DBLP:journals/ior/NilimG05,DBLP:journals/mor/WiesemannKR13,DBLP:journals/mor/XuM12}.
The most explicit game interpretations are given by \cite{DBLP:journals/mor/Iyengar05,DBLP:journals/ior/NilimG05}, which both link reward maximization in $(s,a)$-rectangular RMDPs to turn-based zero-sum stochastic games.
This equivalent game is constructed by adding, in addition to the states $S$, states that correspond with tuples $\tup{s,a}$ of states $s \in S$ and actions $a \in A$.
Then, the agent controls the original states, and nature controls the tuple states.
In a state $s$, the agent chooses an action $a$, upon which the game transitions deterministically to the nature state $\tup{s,a}$.
In this state $\tup{s,a}$, nature chooses a variable assignment.
Given a nature state $\tup{s,a}$ and variable assignment $f\in\cU$, the game transitions stochastically to an agent state $s'$ according to $\cP(f)(s,a,s')$.
The reward function assigns the same value as the reward function of the RMDP in the agent states and zero in nature states.

Changes in the assumptions on nature or the objective require some changes in the translation to stochastic games.
Paper~\cite{DBLP:journals/mor/WiesemannKR13} mentions that a similar construction follows for $s$-rectangular RMDPs, but does not describe the actual game.
As $s$-rectangular RMDPs assume that nature chooses variable assignments without information of the agent's latest action, such games would have to either be partially observable or concurrent.
An average reward objective in RMDPs can be linked to zero-sum mean pay-off games \cite{DBLP:journals/corr/abs-2312-03618}.

A key difference between RMDPs and SGs is that in RMDPs, it is typically assumed that nature plays memoryless, whereas in SGs, both players are allowed to use history.
\cite{DBLP:journals/corr/abs-2312-03618} shows that the assumption of playing against a memoryless nature is nonrestrictive for discounted reward maximization against a convex and compact $s$-rectangular uncertainty set.

\subsection{Robust POMDPs}\label{subsec:rpomdp}
Partially observable MDPs (POMDPs) are an extension of MDPs where it is assumed that the agent cannot directly observe the state.
Instead, the agent receives observations about the state and, optionally, the last action~\cite{DBLP:journals/ai/KaelblingLC98}.
The same extension to the partially observable setting can be made for robust MDPs, resulting in \emph{robust POMDPs} (RPOMDPs).

Computing optimal policies for POMDPs with infinite horizon objectives is undecidable~\cite{DBLP:journals/ai/MadaniHC03}.
Since standard POMDPs are trivially included in RPOMDPs, all decision problems for RPOMDPs are at least as hard as for POMDPs.
Existing approaches for policy computation in RPOMDPs use convex optimization techniques~\cite{DBLP:conf/aaai/Cubuktepe0JMST21,DBLP:conf/ijcai/Suilen0CT20}, robust versions of value iteration~\cite{DBLP:conf/icml/Osogami15}, or recurrent neural networks to learn policies~\cite{galesloot2024pessimistic}.
Other approaches exist but either consider a different notion of optimal policy, such as optimal for one instance in the uncertainty set~\cite{DBLP:journals/ai/ItohN07} or optimal given a pessimism level~\cite{DBLP:journals/jet/Saghafian18}, or have additional assumptions on the uncertainty set, such as uncertainty in the observation function only~\cite{DBLP:conf/cdc/ChamieM18} or existence of a distribution over the uncertainty set~\cite{DBLP:journals/siamjo/NakaoJS21}.
It is shown by~\cite{bovy2024imprecise} that an RPOMDP with an uncertain state-action-based observation function can be transformed to an RPOMDP with a deterministic state-based observation function using a state space expansion.
This transformation allows research to focus on RPOMDPs with uncertainty only in the transition function (and not in the observation function).

Paper~\cite{bovy2024imprecise} defines formal game semantics for RPOMDPs, linking RPOMDPs to turn-based zero-sum partially observable stochastic games (POSGs)~\cite{Springer:HSVI/Delage2023}.
They note that the existing literature on RPOMDPs makes implicit assumptions about uncertainty, which leads to semantically different POSGs and, hence, RPOMDPs with different optimal values.
In particular, \cite{bovy2024imprecise} shows that static and dynamic uncertainty semantics in RPOMDPs no longer coincide, even for $(s,a)$-rectangular uncertainty sets.
For infinite horizon discounted reward maximization,~\cite{DBLP:conf/icml/Osogami15} shows that static and dynamic uncertainty in RPOMDPs do still coincide under $(s,a)$-rectangularity and when nature plays stationary, \ie, without memory.
Finally,~\cite{bovy2024imprecise} shows that nature's ability to observe the agent's last-played action influences the optimal value in both RPOMDPs and RMDPs, see also \Cref{subsec:rmdp:semantics}.

\subsection{Modelling Likelihoods of Transition Functions}\label{subsec:likelihoods}
Thus far, we have discussed models that capture \emph{sets} of possible transition functions and the associated game behaviour of these models.
However, what if certain transition functions are more likely than others?
These likelihoods may result from different experts which value they would assume for, \eg, transition probabilities in a given MDP~\cite{DBLP:conf/iberamia/AndresBMS18}.
The natural question is how we can incorporate this \emph{prior knowledge} about the likelihood of different transition functions.
One option is to neglect the likelihood completely and model the problem as an RMDP.
However, the standard robust analysis of this RMDP (as discussed in \cref{sec:robust:mdps}) may lead to overly conservative results.
In this section, we explore approaches that aim to mitigate this conservatism and rigorously incorporate likelihoods over transition functions.

We can formalize the setting above by modelling the prior knowledge as a probability distribution over the transition function $\cP$ of an RMDP.
Although equivalent, it is often more convenient to model this setting as a pMDP together with a distribution over the parameter values.
As such, these models have been named \emph{uncertain parametric MDPs} (upMDPs) in the literature~\cite{DBLP:journals/sttt/BadingsCJJKT22}.
A common verification question is then to obtain a solution \emph{``that is robust against (for example) at least a $99\%$ probability mass of the distribution.''}
As a concrete application, consider flying a drone in an environment with uncertain weather conditions.
We model this environment as a parametric MDP, where the parameter values are determined by the actual weather conditions.
From historical weather data, we can derive a probability distribution for the weather conditions (and thus the transition probabilities) on a random day.
A natural verification question is then: \emph{``What is the probability that we can safely fly the drone without crashing on a random day?''}

One important question for upMDPs is whether policies can depend on the realization of the (uncertain) parameters.
For the example above, one possible assumption is that we can \emph{first} observe the actual weather conditions, and \emph{then} can compute an optimal policy based on this weather.
This setting has first been investigated by~\cite{DBLP:conf/tacas/Cubuktepe0JKT20} and in more detail by~\cite{DBLP:journals/sttt/BadingsCJJKT22}.
The other possible assumption is that we \emph{cannot} observe the actual weather first and instead need to compute a \emph{single} policy that is robust against all weather conditions.
This setting has been considered by~\cite{DBLP:journals/corr/abs-2312-06344} for upMDPs, and also relates to so-called Bayes-adaptive MDPs which are commonly used in reinforcement learning~\cite{DBLP:conf/nips/GuezSD12,DBLP:conf/aaai/CostenRLH23,DBLP:conf/nips/RigterLH21}.
A variant where parameter values cannot be observed and thus must be learned has been studied by~\cite{DBLP:conf/qest/ArmingBCKS18}.
The latter setting is arguably more difficult to solve due to dependencies between the policies for different weather conditions.
However, which of the two assumptions is more appropriate depends on the context.

In practice, it may be unrealistic to have access to an explicit representation of the distributions over transition functions.
For example, we may have prior knowledge in the form of a finite set of expert demonstrations~\cite{DBLP:conf/aips/PonnambalamOS21}, each of which leads to an MDP with different transition probabilities.
These demonstrations can be interpreted as \emph{samples} from an underlying distribution over the parameters.
This setting has motivated sampling-based verification approaches for upMDPs, such as~\cite{DBLP:conf/tacas/Cubuktepe0JKT20,DBLP:journals/corr/abs-2312-06344}.
A similar setting for continuous-time Markov chains is studied by~\cite{DBLP:conf/cav/BadingsJJSV22,DBLP:journals/corr/abs-2312-06344}.
Generally, these approaches assume access to a finite set of parameter samples and aim to compute a solution with statistical (PAC-style) guarantees on its performance on yet another sample from the underlying distribution.
For example,~\cite{DBLP:journals/sttt/BadingsCJJKT22} obtains PAC guarantees by techniques from \emph{scenario optimization}, which is a methodology to deal with stochastic convex optimization in a data-driven fashion~\cite{DBLP:journals/arc/CampiCG21,DBLP:journals/siamjo/CampiG08}.

\section{Robust MDPs in Practice: Applications and Tools}\label{sec:applications}
In this section, we review two key applications of RMDPs, namely in \emph{learning} and \emph{abstraction} methods, and discuss the current state of tool support.

\subsection{Learning}

A natural application of RMDPs in both formal methods and AI comes in learning MDPs from data.
Naive estimation of the transition probabilities from a finite amount of observations introduces estimation errors.
When following a path through a learned MDP, these errors may accumulate, leading to potentially significant differences in values (and possibly optimal policies) between the learned model and the true underlying MDP~\cite{DBLP:journals/mor/GoyalG23,DBLP:journals/mansci/MannorSST07}.
To account for these errors, confidence intervals around the probabilities or distributions may be computed via, \eg, Hoeffding's inequality~\cite{hoeffding1963probability} or the Weissman bound~\cite{weissman2003inequalities}, and included in the learned model, yielding an RMDP.
Resulting policies and values can then be given a \emph{probably approximately correct} (PAC) guarantee.

Learning MDPs with PAC guarantees has been studied extensively in both formal methods and AI, namely in the form of statistical model checking (SMC) and reinforcement learning (RL).
PAC-SMC is often applied when the original MDP is too large to fit into memory and can mitigate the state-explosion problem (at the cost of precision) when verifying infinite or indefinite horizon objectives.
These methods either build IMDPs by deriving confidence intervals through Hoeffding's inequality or construct lower and upper Bellman equations that are updated directly, implicitly performing robust dynamic programming~\cite{DBLP:conf/cav/AshokKW19,DBLP:conf/tacas/DacaHKP16}.

In contrast, RL is primarily concerned with finding an optimal policy that maximizes discounted or finite horizon reward objectives through efficient exploration~\cite{DBLP:books/lib/SuttonB98}.
RMDPs, and specifically \Lone-MDPs based on the aforementioned Weissman bound, are used to achieve PAC guarantees on the learned model~\cite{DBLP:journals/jmlr/StrehlLL09,DBLP:journals/jcss/StrehlL08} or efficient exploration through optimistic policies~\cite{DBLP:journals/jmlr/JakschOA10}.

It should be noted that the PAC-MDP framework of~\cite{DBLP:journals/jmlr/StrehlLL09} explicitly requires the sample efficiency of a learning algorithm to be polynomial in the input to be considered (efficiently) PAC.
As a consequence, any non-finitary objective is not PAC-learnable following the PAC-MDP framework, and PAC-SMC methods are said to give anytime or best-effort guarantees~\cite{DBLP:conf/ijcai/YangLC22}.
Recent work investigates techniques to reduce the amount of data required to achieve PAC guarantees in both SMC and RL~\cite{DBLP:journals/corr/abs-2404-05424,DBLP:conf/ijcai/WienhoftSSDB023}.

Two subfields of RL that also commonly use RMDPs are the offline RL problem of \emph{safe policy improvement} (SPI) and \emph{robust RL}. 
In SPI, only a previously collected data set and the \emph{behaviour} policy that collected it are given, and no further data collection is allowed.
The SPI problem is to compute a new policy that outperforms the behaviour policy with a PAC-style guarantee.
Solutions to this problem often construct (implicit) \Lone-MDPs~\cite{DBLP:conf/nips/GhavamzadehPC16,DBLP:conf/icml/LarocheTC19,DBLP:conf/aaai/SimaoS023}. 
Robust RL is a broad field that considers RL under various kinds of uncertainty, disturbances, or structural changes perturbations, such as non-stationary environments~\cite{DBLP:conf/nips/SuilenS0022}.
We refer to the following recent survey for more on robust RL~\cite{DBLP:journals/make/MoosHASCP22}.

\subsubsection{Aleatoric and epistemic uncertainty.}
Uncertainty that may be reduced by collecting more data, as encountered in these learning settings, is commonly referred to as \emph{epistemic uncertainty}~\cite{DBLP:journals/sttt/BadingsSSJ23,DBLP:journals/ml/HullermeierW21}. 
Uncertainty that cannot be reduced but is known to be inherent to the system, such as the probability distributions in a standard MDP, is called \emph{aleatoric uncertainty}.
We emphasize that an RMDP's uncertainty set is not necessarily epistemic and that whether the uncertainty may be reduced by collecting more data is an additional assumption about the specific scenario in which the RMDP is used.

\subsection{Abstraction}
RMDPs are commonly used to model abstractions of more complex systems.
The general idea is that states of a (non-robust) MDP can be aggregated by overapproximating the transition probabilities in the uncertainty set of an RMDP~\cite{DBLP:conf/isola/Jaeger0BLJ20}.
This idea at least dates back to~\cite{DBLP:journals/iandc/LarsenS91} and has already been identified as an interesting application of RMDPs %
in~\cite{DBLP:journals/ai/GivanLD00}.
Since then, such abstraction techniques have been used across areas, including formal methods, control theory, and AI.

First of all, game-based abstraction of MDPs in the form of IMDPs has been studied by~\cite{DBLP:journals/fmsd/KattenbeltKNP10,DBLP:conf/qest/KwiatkowskaNP06}.
So-called 3-valued abstractions of Markov chains are developed by~\cite{DBLP:conf/spin/FecherLW06}, who abstract the system into an interval Markov chain whose labelling function has three possible values (true, false, or don't know).
A similar approach for 3-valued abstraction of CTMCs is presented by~\cite{DBLP:conf/cav/KatoenKLW07}.
Probabilistic bisimulation of IMDPs to reduce the number of states is considered by~\cite{DBLP:conf/lata/HashemiH0STW16}.

In control theory, models typically have continuous state and action spaces.
A popular approach to synthesizing provably correct control policies is to generate a finite-state abstraction of the continuous model~\cite{Abate2008probabilisticSystems,Alur2000,DBLP:journals/tac/LahijanianAB15}.
Under an appropriate simulation relation, satisfaction guarantees of temporal logic formulae carry over from the abstract to the continuous model~\cite{DBLP:journals/tac/GirardP07}.
Various papers generate IMDP abstractions of stochastic systems~\cite{DBLP:conf/aaai/BadingsRA023,DBLP:journals/jair/BadingsRAPPSJ23,DBLP:journals/corr/abs-2404-08344,DBLP:journals/csysl/LavaeiSFZ23}, and tool support has been developed by, \eg,~\cite{DBLP:conf/tacas/CauchiA19,DBLP:journals/corr/abs-2401-03555}.
Similar to game-based abstraction, the general idea is to use the probability intervals to capture uncertainties and abstraction errors.
For further details on abstractions in control, we refer to the survey~\cite{DBLP:journals/automatica/LavaeiSAZ22}.

\subsection{Tool Support}\label{subsec:tools}
General-purpose tool support for RMDPs is still relatively limited compared to other models.
The probabilistic model checkers \Prism~\cite{DBLP:conf/cav/KwiatkowskaNP11} and \Storm~\cite{DBLP:journals/sttt/HenselJKQV22} both support basic IMDP model checking, with \Prism's provision slightly more advanced, \eg, in terms of user support for modelling.
\Storm, on the other hand, has more advanced support for pMDPs, and has been used as a back-end in several of the previously discussed works, \eg,~\cite{DBLP:journals/sttt/BadingsCJJKT22,DBLP:conf/cav/BadingsJJSV22,DBLP:journals/tac/CubuktepeJJKT22,DBLP:conf/ijcai/Suilen0CT20}.
Other IMDP tools have also begun to be developed, such as the Julia-based \textsc{IntervalMDP.jl}~\cite{MML24}.
\section{Robust MDPs in the Future}\label{sec:challenges}

We conclude this survey with some currently active directions for the development and application of RMDPs.

\subsection*{Tools, Benchmarks, and Evaluation}
As mentioned in \Cref{subsec:tools}, tool support for RMDPs is still young, with a particular focus on IMDPs. 
Extending support to \Lone-MDPs, a larger range of objectives including discounted reward and temporal logic objectives, and possibly $s$-rectangular RMDPs, would give further impulse to the development of new techniques exploiting the theory of RMDPs in, \eg, learning or abstraction.

Alongside tools, we also see a clear need for a rich benchmark set to evaluate and compare RMDP algorithms and tools.
Initiatives such as a rich comparison of algorithms~\cite{DBLP:conf/tacas/HartmannsJQW23} or extensive (tool) benchmarking on a standardized set of models~\cite{andriushchenko2024a,DBLP:conf/tacas/HartmannsKPQR19} could shed light on the differences between various methods in theory and practice.

Notably, and contrary to what has been experimentally verified for MDPs~\cite{DBLP:conf/tacas/HartmannsJQW23}, robust policy iteration seems to be the preferred way to solve RMDPs~\cite{DBLP:journals/jmlr/HoPW21,DBLP:journals/informs/KaufmanS13}.
The argument is that it performs fewer inner minimization problems compared to robust value iteration, which, depending on the shape of the uncertainty set, may become computationally expensive.
An extensive experimental evaluation confirming this hypothesis is, to the best of our knowledge, missing as of yet.

\subsection*{Algorithmic and Theoretical Advances in Multi-Environment MDPs}
In \Cref{subsec:rmdp:instances}, we already identified MEMDPs as an interesting class of RMDPs as they naturally model a finite set of possible MDPs and may also be viewed as POMDPs with additional structure.
Complexity results on almost-sure objectives show how this additional structure can be exploited to build computationally more efficient algorithms than their (general) POMDP counterpart~\cite{DBLP:conf/tacas/VegtJJ23}.
We believe this direction of research should be continued and further investigated for quantitative objectives such as expected reward.

\subsection*{Uncertainty Assumptions in RMDPs}
While the type of rectangularity assumptions about the uncertainty set are often made explicit, other assumptions, such as static and dynamic uncertainty and the type of nature policies considered, are mostly left implicit.
In various cases, it is known that these assumptions do not influence the optimal policies and values and can, therefore, be ignored~\cite{DBLP:journals/corr/abs-2312-03618,DBLP:journals/mor/Iyengar05}, as also discussed in section \Cref{sec:robust:mdps,sec:connections}.
However, many combinations of assumptions are yet to be investigated.
\cite{bovy2024imprecise} shows cases where assumptions on the uncertainty set influence the optimal policy and value for reward maximization in RPOMDPs.
Therefore, richer (temporal logic) objectives for RMDPs may likely be subject to a similar influence.
We believe expanding the knowledge of whether and when assumptions on the RMDP influence the optimality of policies and values is an interesting and necessary direction for future research.

\section{Conclusion}
We presented a short survey on robust MDPs, from the basic foundations, including semantics and solution methods, to common applications in formal methods and AI.
We discussed RMDP semantics and how dynamic programming can be extended to \emph{robust} dynamic programming in the $(s,a)$-rectangular case.
Finally, we summarized the current and (possibly) future position RMDPs take within the formal methods and AI communities in terms of connections with other models and applications.

\section*{Acknowledgements}
This work was supported by the ERC Starting Grant 101077178
(DEUCE) and the European Union’s
Horizon 2020 research and innovation programme (FUN2MODEL, grant agreement No. 834115), as well as the NWO grants\\
OCENW.KLEIN.187 and NWA.1160.18.238 (PrimaVera).

\bibliographystyle{splncs04}
\bibliography{references}

\end{document}